# On the Identifiability of the Post-Nonlinear Causal Model

**Kun Zhang**
Dept. of Computer Science and HIIT
University of Helsinki
Finland

**Aapo Hyvärinen**
Dept. of Computer Science, HIIT
and Dept. of Mathematics and Statistics
University of Helsinki
Finland

## Abstract

By taking into account the nonlinear effect of the cause, the inner noise effect, and the measurement distortion effect in the observed variables, the post-nonlinear (PNL) causal model has demonstrated its excellent performance in distinguishing the cause from effect. However, its identifiability has not been properly addressed, and how to apply it in the case of more than two variables is also a problem. In this paper, we conduct a systematic investigation on its identifiability in the two-variable case. We show that this model is identifiable in most cases; by enumerating all possible situations in which the model is not identifiable, we provide sufficient conditions for its identifiability. Simulations are given to support the theoretical results. Moreover, in the case of more than two variables, we show that the whole causal structure can be found by applying the PNL causal model to each structure in the Markov equivalent class and testing if the disturbance is independent of the direct causes for each variable. In this way the exhaustive search over all possible causal structures is avoided.

## 1   INTRODUCTION

Traditionally, causal discovery algorithms, which may be constraint-based or score-based, produce a Markov equivalent class of the causal models, in which some causal directions may be undetermined (Spirtes et al., 2001; Pearl, 2000). On the other hand, a functional causal model, which expresses each variable as a function of its direct causes and the independent disturbance (Pearl, 2000), if well specified, can explain the data generating process and help find all causal relations among the variables. For example, under the condition that causal relations are linear and acyclic, the non-Gaussianity of the disturbances could help find the whole causal model uniquely (Shimizu et al., 2006), by resorting to the independent component analysis (ICA) technique (Hyvärinen et al., 2001). However, if the assumed functional causal model is not capable of approximating the true data generating process, the results may be misleading. Therefore, if the specific knowledge about the data generating mechanism is not available, to make it useful in practice, the assumed causal model should be general enough, such that it can reveal the data generating processes approximately; at the same time, the model should be identifiable, i.e., it is asymmetrical in causes and effects and is capable of distinguishing between them.

Although the linearity assumption greatly simplifies causal analysis, in some situations nonlinear effects in the system are not negligible. In particular, the recently proposed post-nonlinear (PNL) causal model takes into account the nonlinear effect of the causes, the noise effect, and sensor or measurement nonlinear distortion in the observed variables (Zhang & Hyvärinen, 2008). Mathematically, with the causal structure represented by a directed acyclic graph (DAG), it expresses each variable $x_i$ as

$$x_i = f_{i,2}(f_{i,1}(pa_i) + e_i), \quad i = 1, ..., n, \qquad (1)$$

where $pa_i$ contains the direct causes of $x_i$, $f_{i,1}$ denotes the nonlinear effect of the causes, $e_i$ is the independent disturbance, and $f_{i,2}$ denotes invertible post-nonlinear distortion in variable $x_i$. It includes the so-called additive noise model (Hoyer et al., 2009) as a special case in which the nonlinear distortion $f_{i,2}$ does not exist. It is also related to the idea of causal reasoning based on evaluating the complexity of conditional densities (Sun et al., 2008), since it intrinsically admits a simple expression for the conditional density of the effect given the causes. This model has been used to distinguish between causes and effects for the "Cause-effect pairs" task (Mooij et al., 2008) in the second causality challenge, and gave clearly the best results (the identified causal directions are correct for all eight data sets). Its good performance is partially due to the allowance of the measurement distortion $f_{i,2}$, which is frequently encountered in practice.

Despite its success in solving some real-world problems, there are two unsolved problems related to the PNL causal model, and they are addressed in this paper. One is the identifiability, a crucial issue, of this model. Although it was supported by empirical results and was touched in Zhang and Hyvärinen (2008), it is far from complete. Here we give a systematical investigation of its identifiability in the two-variable case, under the assumption that the density of the disturbance has an unbounded support. We show that the model is generally identifiable, and give all the non-identifiable cases, some of which are illustrated by simulation studies. Our results also have some by-products. Previously, Zhang and Hyvärinen (2008) investigated this issue, by relating this model to the PNL mixing ICA problem (Achard & Jutten, 2005), under the constraint that the nonlinear effect of the cause, $f_{i,1}$, is invertible. Our findings reveal that their results are not precise, and further provide counterexamples to the separability theorem of the PNL mixing ICA problem reported in Achard and Jutten (2005), which their results are based on. This finding may also be of interest to the ICA community, since PNL mixing ICA is an important and widely used nonlinear ICA model. In addition, our results on the identifiability of the PNL causal model also apply to the additive noise model (Hoyer et al., 2009), since the latter model is a special case of the former one.

The other problem is how to find the causal relations among more than two variables implied by this model. One may search all possible causal structures and test if they are consistent with the data in a brute-force way, but it involves high computational load and becomes impractical as the variable number increases. We provide some fundamental results for this issue. It is shown that causal discovery based on the PNL causal model for more than two variables can be achieved in two steps: after obtaining the equivalent class, one can identify the undetermined causal relations by applying this model to the causal structures in the equivalent class and testing if each disturbance is independent of the parents associated with the same variable. Consequently, the search space is greatly reduced, and statistical tests of mutual independence between more than two variables are avoided.

## 2 IDENTIFIABILITY

In this section we focus on the two-variable case. We investigate the identifiability of the PNL causal model, in particular, its direction, by a proof by contradiction.

We assume the causal model holds in both directions $x_1 \rightarrow x_2$ and $x_2 \rightarrow x_1$, and show that this implies some very strong conditions on the distributions and functions involved in the model.

Assume that the data $(x_1, x_2)$ are generated by the post-nonlinear (PNL) causal model with the the causal relation $x_1 \rightarrow x_2$. This data generating process can be described as

$$x_2 = f_2(f_1(x_1) + e_2), \tag{2}$$

where $x_1$ and $e_2$ are independent, function $f_1$ is non-constant, and $f_2$ is invertible. If the other causal direction, $x_2 \rightarrow x_1$ is true, the data generating process given by the PNL causal model is

$$x_1 = g_2(g_1(x_2) + e_1), \tag{3}$$

where $x_2$ and $e_1$ are independent, $g_1$ is non-constant, and $g_2$ is invertible.

**Notation.** The following notations are used hereafter. Suppose that both (2) and (3) hold. Random variables $t_1$ and $z_2$ and functions $h$ and $h_1$ defined as follows:

$$t_1 \triangleq g_2^{-1}(x_1), \qquad z_2 \triangleq f_2^{-1}(x_2),$$
$$h \triangleq f_1 \circ g_2, \qquad h_1 \triangleq g_1 \circ f_2.$$

That is, $h(t_1) = f_1(g_2(t_1)) = f_1(x_1)$, and similarly, $h_1$ is a function of $z_2$. Moreover, we let $\eta_1(t_1) \triangleq \log p_{t_1}(t_1)$, and $\eta_2(e_2) \triangleq \log p_{e_2}(e_2)$.[1] The following theorem gives the constraint that $p_{t_1}$, $p_{e_2}$, and $h$ must satisfy to make both (2) and (3) hold.

**Theorem 1** *Assume that $(x_1, x_2)$ can be described by both of the causal relations given in (2) and in (3). Further suppose that involved densities and nonlinear functions $p_{t_1}$, $p_{e_2}$, $f_1$, $f_2$, $g_1$, and $g_2$ are third-order differentiable, and that $p_{e_2}$ is positive on $(-\infty, +\infty)$. We then have the following equation for every $(x_1, x_2)$ satisfying $\eta_2'' h' \neq 0$:*

$$\eta_1''' - \frac{\eta_1'' h''}{h'} = \left( \frac{\eta_2' \eta_2'''}{\eta_2''} - 2\eta_2'' \right) \cdot h' h'' - \frac{\eta_2'''}{\eta_2''} \cdot h' \eta_1'' + \eta_2' \cdot \left( h''' - \frac{h''^2}{h'} \right), \tag{4}$$

*and $h_1$ depends on $\eta_1$, $\eta_2$, and $h$ in the following way:*

$$\frac{1}{h_1'} = \frac{\eta_1'' + \eta_2' h'^2 - \eta_2' h''}{\eta_2'' h'}. \tag{5}$$

See Appendix for its proof. In the special case with $f_2$ and $g_2$ being the identity mapping, the above theorem becomes Theorem 1 in Hoyer et al. (2009), which was given for the additive noise model. In general, it is not obvious if Theorem 1 holds for a practical problem. Therefore, below we provide conditions which are easy to verify, by investigating (4) in Theorem 1. To this end, we first give some definitions and lemmas.

---

[1] For the sake of conciseness, sometimes we drop the arguments of the functions in the presentation.

**Definition 2** *The density of the continuous variable $v$, denoted by $p_v$, is said to be log-mixed-linear-and-exponential (log-mix-lin-exp) if it is of the form $\log p_v = c_1 e^{c_2 v} + c_3 v + c_4$, where $c_1$, $c_2$, $c_3$, and $c_4$ are constants. Clearly, to make $p_v$ a valid density, we have $c_1 < 0$ and $c_2 c_3 > 0$. This type of distributions includes the Type-1 Gumbel distribution as a special case when $c_2 = c_3$.*

**Definition 3** *The density $p_v$ is said to be a generalized mixture of two exponentials if $p_v \propto (c_1 e^{c_2 v} + c_3 e^{c_4 v})^{c_5}$, with constants $c_i$, or equivalently, if $\log p_v = d_1 v + d_2 \log(d_3 + d_4 e^{d_5 v}) + d_6$, with constant $d_i$.*

**Definition 4** *The density $p_v$ is said to be one-sided asymptotically exponential, if $(\log p_v)' \to c$, as $v \to -\infty$ or as $v \to +\infty$, where $c$ is a non-zero constant. It is said to be two-sided asymptotically exponential if $(\log p_v)' \to c_1$, as $v \to -\infty$, and $(\log p_v)' \to c_2$, as $v \to +\infty$, where $c_1$ and $c_2$ are non-zero constants.*

Clearly, log-mix-lin-exp densities are special cases of the one-sided asymptotically exponential densities, and two-sided asymptotically exponential densities include those generalized mixtures of two exponentials. Now we give a lemma to characterize some properties of a density function or its logarithm.

**Lemma 5** *Suppose the density $p_v$ is differentiable on $(-\infty, +\infty)$ and monotonic for sufficiently large $v$. We have the following. (i) If $p_v$ is positive on $(-\infty, +\infty)$, then $\log p_v \to -\infty$ as $|v| \to +\infty$. (ii) If there only exists one point, denoted by $v = C$, satisfying $(\log p_v)' = 0$, then on the support of $p_v$, $(\log p_v)' > 0$ for $v < C$, and $(\log p_v)' < 0$ for $v > C$.*

This lemma is obvious and proof is skipped. The following lemmas discuss some special and simple solutions to (4) in Theorem 1.

**Lemma 6** *Under the assumptions made in Theorem 1, if $e_2$ is Gaussian, then $h$ is linear and $t_1$ is also Gaussian.*

**Lemma 7** *Under the assumptions made in Theorem 1, if function $h = f_1 \circ g_2$ is linear, one of the following must be true:*

  *i. $t_1$ and $e_2$ are both Gaussian, and $h_1$ is also linear.*

  *ii. The densities of $t_1$ and $e_2$ are log-mix-lin-exp, $h'_1(z_2)$ is strictly monotonic, and $h'_1(z_2) \to 0$, as $z_2 \to +\infty$ or as $z_2 \to -\infty$.*

For proofs, see Appendix. We now give Theorem 8 to consider all the situations in which the PNL causal model is not identifiable.

**Theorem 8** *Suppose that $\eta''_2 h' \neq 0$ at every point or that it is zero at only some discrete points. Under the assumptions made in Theorem 1, we have that $p_{e_2}$ and $p_{t_1}$, as well as $h$, must satisfy one of the five conditions listed in Table 1.*[2]

The proof is rather long and complicated, so we just give the basic idea and its outline; see Appendix. When the nonlinear distortions $f_2$ and $g_2$ are constrained to be the identity mapping, this result also applies to the identifiability of the additive noise model. We give the following remarks on the situations listed in Table 1. Situation I in Table 1 is the linear Gaussian case, which is well know to be not identifiable. Situations II~V are novel. As discussed in Zhang and Hyvärinen (2008), if $f_1$ is constrained to be invertible, the PNL causal model (2) can be transformed to the PNL mixing ICA model (Achard & Jutten, 2005), with $f_1(x_1)$ and $e_2$ considered as sources. The identifiability of the former model is then implied by the separability of the latter one. Previously, under weak conditions, it was shown that the PNL mixing ICA model is separable if at most one of the sources is Gaussian (Achard & Jutten, 2005). In Situations II~V, $p_{e_2}$ is not Gaussian, but the causal model is not identifiable, and consequently the corresponding PNL mixing ICA model is not separable. This means that the established separability results of the PNL mixing ICA problem have some flaws and require further investigation.

Below we give some corollaries which follow from Theorem 8. They can be easily exploited to examine if the causal relation between two variables is unique. The following assumptions are made in the corollaries.

**A1.** The data $(x_1, x_2)$ are generated by the PNL causal model (2), with $f_1$ and $f_2$ being third-order differentiable.

**A2.** Densities $p_{e_2}$ and $p_{x_1}$ are third-order differentiable, $p_{e_2}$ is positive on $(-\infty, +\infty)$, and $\eta''_2 = (\log p_{e_2})''$ is zero at most at some discrete points.

The following corollary immediately follows Theorem 8, so its proof is skipped.

---
[2] Note that the identifiability of the PNL causal model (2) depends directly on the distribution of $t_1$ and $e_2$, instead of the distribution of the observed variables $x_1$ and $x_2$.

Table 1: All situations in which the PNL causal model is not identifiable.

|   | $p_{e_2}$ | $p_{t_1}$ $(t_1 = g_2^{-1}(x_1))$ | $h = f_1 \circ g_2$ | Remark |
|---|---|---|---|---|
| I | Gaussian | Gaussian | linear | $h_1$ also linear |
| II | log-mix-lin-exp | log-mix-lin-exp | linear | $h_1$ strictly monotonic, and $h_1' \to$ 0, as $z_2 \to +\infty$ or as $z_2 \to -\infty$ |
| III | log-mix-lin-exp | one-sided asymptotically exponential (but not log-mix-lin-exp) | $h$ strictly monotonic, and $h' \to 0$, as $t_1 \to +\infty$ or as $t_1 \to -\infty$ | — |
| IV | log-mix-lin-exp | generalized mixture of two exponentials | Same as above | — |
| V | generalized mixture of two exponentials | two-sided asymptotically exponential | Same as above | — |

**Corollary 9** *Suppose that assumptions A1 & A2 hold. If $p_{e_2}$, the density of the disturbance, is not Gaussian, nor log-mix-lin-exp, nor a generalized mixture of two exponentials, then the PNL causal model (2) is identifiable.*

Corollary 10 considers the identifiability of the PNL causal model with $f_1$, the nonlinear effect of the cause, being non-invertible.

**Corollary 10** *Suppose that assumptions A1 & A2 hold. If function $f_1$ is not invertible, then the PNL causal model (2) is identifiable.*

For proof, see Appendix. This result is intuitively appealing, and confirms the finding in Friedman and Nachman (2000).

## 3 NONLINEAR ICA-BASED IDENTIFICATION METHOD

If data $(x_1, x_2)$ follow the PNL causal model with causal direction $x_1 \to x_2$, from (2), we can see that the disturbance $e_2$, which is independent from $x_1$, can be expressed in terms of $x_1$ and $x_2$:

$$e_2 = f_2^{-1}(x_2) - f_1(x_1).$$

This provides a way to verify if $x_1 \to x_2$ holds according to the PNL causal model, as given below.

Under the hypothesis $x_1 \to x_2$, we can estimate the disturbance $e_2$ by finding functions $l_1$ and $l_2$ such that $\hat{e}_2 = l_2(x_2) - l_1(x_1)$ is independent from $x_1$. This is then a constrained nonlinear ICA problem, and can be achieved by minimizing $I(x_1, \hat{e}_2)$, the mutual information between $x_1$ and $\hat{e}_2$ (Zhang & Hyvärinen, 2008). After some simplifications, one can see $I(x_1, \hat{e}_2) = -E \log p_{\hat{e}_2}(\hat{e}_2) - E \log |l_2'(x_2)| + H(x_1) - H(\hat{e}_2)$. As the last two terms do not depends on $l_1(x_1)$ and $l_2(x_2)$, minimizing $I(x_1, \hat{e}_2)$ is equivalent to maximizing $E \log p_{\hat{e}_2}(\hat{e}_2) + E \log |l_2'(x_2)|$. Following Zhang and Hyvärinen (2008), we use multi-layer perceptrons

(MLP's) to represent $l_1$ and $l_2$. The involved parameters can then be learned by gradient-based methods. Finally, if statistical independence tests, such as the kernel-based test (Gretton et al., 2008), confirm that $\hat{e}_2$ is independent from $x_1$, $x_1 \to x_2$ is supported by the PNL causal model, and the learned $l_1$ and $l_2$ provide an estimate of $f_1$ and $f_2^{-1}$, respectively.

To find the causal relation between $x_1$ and $x_2$ implied by the PNL causal model, one needs to test both hypotheses $x_1 \to x_2$ and $x_2 \to x_1$, using the method just described. If exactly one of them holds, the causal relation between $x_1$ and $x_2$ implied by the PNL model has been successfully found. If neither of them holds, there is no PNL causal relation between $x_1$ and $x_2$. If both hold, the cause and effect could not be distinguished by the PNL causal model; additional information, such as the smoothness of the involved nonlinearities, may help find the causal model with a lower complexity.

## 4 MORE THAN TWO VARIABLES

The PNL acyclic causal model (1) is applicable in the case of more than two variables. When there are only very few variables, one may use a brute-force search to find the causal relations, like the estimation of the additive noise model in Hoyer et al. (2009); for each possible acyclic causal structure, represented by a DAG, we use the nonlinear ICA-based approach to estimate the corresponding disturbances, and then verify if they are mutually independent by performing independence tests. The simplest causal model which gives independent disturbances is preferred. Clearly this approach may encounter two difficulties. One is that the test of mutual independence is difficult to do when we have many variables. The other is that the search space of all possible DAG's increases too rapidly with the variable number. Consequently, this approach involves high computational load, and does not scale well with the number of variables.

A more practical approach to finding the causal relations implied by the PNL causal model consists of the following two steps. We first use conditional independence-based methods to find the d-separation equivalent class. Next, the PNL causal model is used to identify the causal directions that cannot be determined in the first step: for each causal structure contained in the equivalent class, we estimate the disturbances, and determine if this causal structure is plausible, by examining if the disturbance in a variable $x_i$ is independent from the parents of $x_i$. Consequently, one avoids the exhaustive search over all possible causal structures and high-dimensional statistical tests of mutual independence. The validity of this approach is supported by the following theorem.

**Theorem 11** *When fitting variables $x_1, ..., x_n$ to the PNL acyclic causal model (1) with the causal structure represented by the DAG $\mathcal{G}$, the disturbances $e_i$ are mutually independent if and only if the causal Markov condition holds (i.e., each variable $x_i$ is independent of its non-descendants conditional on its parents $pa_i$ in $\mathcal{G}$), and the disturbance $e_i$ in $x_i$ is independent of the parents of $x_i$.*

Proof is given in Appendix. An important issue in this approach is how to perform conditional independence tests for variables with nonlinear causal relations. Generally speaking, the results of nonparametric conditional independence tests may be unreliable when the conditional set contains many variables, due to the curse of dimensionality. Traditionally, in the implementation of most conditional independence-based causal discovery algorithms, such as PC (Spirtes et al., 2001) and IC (Verma & Pearl, 1990), it is assumed that the variables are either discrete or Gaussian with linear causal relations. Although this assumption greatly simplifies the difficulty in conditional independent tests, it usually does not hold in our case which involves nonlinear causal effects and non-Gaussian variables. Alternatively, in our case, one may simplify the conditional independence test procedure by making use of the particular structure of the PNL causal model. This is out of the scope of this paper and left for future work.

# 5    SIMULATIONS

The PNL causal model has been applied for causal discovery of two variables with real-world data in Zhang and Hyvärinen (2008). It successfully identified the causal directions for all eight data sets in the "Cause-effect pairs" task (Mooij et al., 2008) included in Causality Challenge #2, without any background knowledge. This demonstrated the practical usefulness of this model and the identification method for some real-world problems. Here we conduct simulations to verify the non-identifiable conditions of this model with two variables, given in Theorem 8. In particular, Situation I in Table 1 is well known to be not identifiable, but the others are completely new and interesting. Here we use illustrative examples to show the non-identifiability in Situations II and V.

## 5.1    ON SITUATION II IN TABLE 1

In Situation II in Table 1, function $h = f_1 \circ g_2$ is linear and the densities of $t_1$ and $e_2$ are log-mix-lin-exp (given by (10) and (9) in Appendix, respectively), and the PNL causal model is not identifiable. To illustrate that, one just needs to confirm that $t_1$ and $z_2$ are independent of $e_2$ and $e_1$, respectively; independence between $t_1$ and $e_2$ means that the causal relation $x_1 \to x_2$ holds, as described by (2); in addition, if $z_2$ and $e_1$ given by (6) and (7) (see Appendix) are also independent, one can see that $x_2 \to x_1$ also holds, and that the PNL model is not identifiable.

Given $c_1$, $c_2$, $c_3$, and $c_5$ in (10) and (9), one can find the constants $c_4$ and $c_7$ such that $p_{t_1} = e^{\eta_1}$ and $p_{e_2} = e^{\eta_2}$ are valid densities. We then generate random numbers as realizations of the independent variables $t_1$ and $e_2$, by applying the inverse of their cumulative distribution functions (CDF's) to independent and uniformly distributed variables. According to (7), variable $z_2$ involved in the causal relation $x_2 \to x_1$ is calculated as $z_2 = e_2 + h(t_1)$. The other variable, $e_1$, can be found according to (6), if $h_1$ is known. According to (11), we have

$$h_1' = e^{c_1 z_2 - c_5 + c_2} \Big/ \Big( \frac{1}{h'} + h' e^{c_1 z_2 - c_5 + c_2} \Big).$$

This yields $h_1 = \frac{1}{h' c_1} \log |\frac{1}{h'} + h' e^{c_1 z_2 - c_5 + c_2}| + c_8$, where $c_8$ is an arbitrary constant. Substituting $h_1$ into (6), we can find $e_1$.

In our simulation study, we set $h(t_1) = -t_1$, $c_1 = 0.3$, $c_2 = -1$, $c_3 = 1$, $c_5 = -1$ $c_8 = 0$, and variables $t_1$ and $e_2$ were made zero-mean. 2000 samples were drawn. Fig. 1 (a) gives the scatter plot of $t_1$ and $e_2$, as well as their marginal histograms. Finally, the scatter plot of $z_2$ and $e_1$ is shown in Fig. 1 (b). To verify if $z_2$ and $e_1$ are independent, we conducted the kernel-based statistical independence tests (Gretton et al., 2008). The left part of Table 2 presents the independence test results for the pairs $(t_1, e_2)$ and $(z_2, e_1)$. For both pairs, The independence hypothesis is accepted at the significance level $\alpha = 0.05$. That is, in this situation, the PNL causal model cannot distinguish the cause from effect, if we do not have further knowledge about the data generating process.

We further generated observed data $(x_1, x_2)$ from variables $t_1$ and $e_2$, aiming to verify if the nonlinear ICA-based identification method can detect both causal directions. Note that the identifiability of the PNL causal model only depends the distributions of $t_1$ and $e_2$ and function $h(t_1) = f_1(g_2(t_1))$. In situation II in Table 1, $f_1$, $f_2$, and $g_2$ can be arbitrary, given that $h = f_1 \circ g_2$ satisfies the condition.[3] In this simulation, we let $g_2(t_1) = t_1/2 + t_1^{1/3}$; since $h = f_1 \circ g_2$, $f_1$ was then constructed as $f_1 = h \circ g_2^{-1}$. we let $f_2(h(t_1) + e_2) = \tanh((h(t_1)+e_2)/8)$. Finally, $(x_1, x_2)$ were constructed from $t_1$ and $e_2$ by $x_1 = g_2(t_1)$ and (2). We applied the nonlinear ICA-based method given in Section 3, and obtained $\hat{e}_2$ and $\hat{e}_1$, which are the estimate of the disturbance under the hypotheses $x_1 \rightarrow x_2$ and $x_2 \rightarrow x_1$, respectively. The right part of Table 2 reports the results of independence tests for the pairs $(x_1, \hat{e}_2)$ and $(x_2, \hat{e}_2)$. Both pairs accept the independence hypothesis, meaning that both the estimated PNL causal model with $x_1 \rightarrow x_2$ and that with $x_2 \rightarrow x_1$ can explain the data.

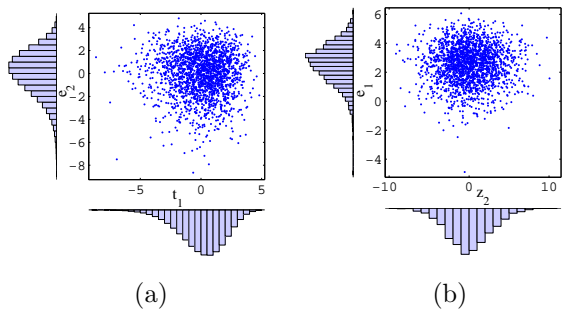

(a)                              (b)

Figure 1: (a) Scatter plot of $t_1 = g_2^{-1}(x_1)$ and $e_2$ in Simulation 1. (b) That of $z_2 = f_2^{-1}(x_2)$ and $e_1$.

Table 2: Result of kernel-based independence tests in Simulation 1, at the significance level $\alpha = 0.05$. The independence hypothesis is accepted in all the four cases. See main text for further explanations.

|           | $t_1\&e_2$ | $z_2\&e_1$ | $x_1\&\hat{e}_2$ | $x_2\&\hat{e}_1$ |
|-----------|--------|--------|--------|--------|
| Threshold | 0.5669 | 0.5665 | 0.4347 | 0.5778 |
| Statistic | 0.3497 | 0.1226 | 0.1430 | 0.1842 |

## 5.2 ON SITUATION V IN TABLE 1

We next verify Situation V in Table 1. In the proof of Theorem 8, Solution 1 to the functional equation (12), given by (14), with the constraints $A_3 = -A_2$ and $A_4 \neq 0$, results in this situation. Generally speaking, in this situation, the solutions of $\eta_1$ and $h$ cannot be expressed in terms of elementary functions. We then found numerical solutions of $\eta_1'$, $h'$, and $\eta_2'$ to the overdetermined system of ordinary differential equations (ODE's) given in (14) using MATLAB. We let $A_1 = 2$, $A_2 = 0.4$, $A_4 = 1$, $c_6 = -10$, $c_7 = 1.6$, and the initial conditions are $\eta_1'(t_1 = 0) = 0$, $\eta_2'(e_2 = 0) = 0$, $h'(t_1 = 0) = 3$. The solutions of $\eta_1'$, $h'$, and $\eta_2'$ are depicted in Fig. 2(a). One can see that $h'$ is clearly close to zero for large $t_1$, consistent with Situation V. Furthermore, we set $h(t_1 = 0) = 0$, and obtained the solutions of $h$; also noting that the total integral of $p_{t_1} = e^{\eta_1^0}$ and $p_{e_2} = e^{\eta_2}$ is one, finally we obtained the solutions of $\eta_1$, $h$, and $\eta_2$, as plotted in Fig. 2(b).

---
[3]Note that if $f_1$ is not invertible, the causal model is always identifiable; see Corollary 10.

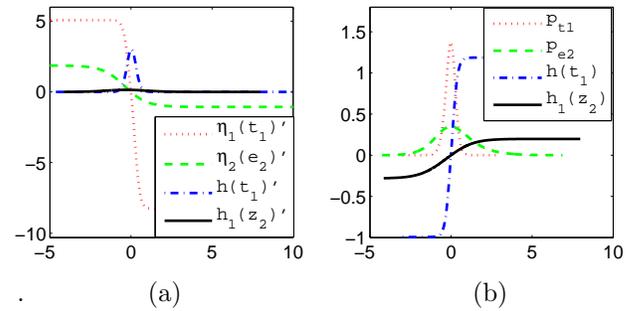

(a)                              (b)

Figure 2: An example of Situation V in Table 1. (a) The curves of $\eta_1(t_1)'$, $\eta_2(e_2)'$, $h(t_1)'$, and $h_1(z_2)'$. $\eta_1(t_1)'$, $\eta_2(e_2)'$, and $h(t_1)'$ were obtained by solving the system of ODE's in (14) numerically. $h_1(t_1)'$ was given by (5). (b) The curves of $p_{t_1} = e^{\eta_1}$, $p_{e_2} = e^{\eta_2}$, $h(t_1)$, and $h_1(z_2)$. See main text for detailed explanations.

We used radial basis networks (RBF's) to learn the probability density functions (PDF's) of $t_1$ and $e_2$ from their numerical values, and then drew random samples of $t_1$ and $e_2$ by applying the inverse of their CDF's to independent and uniformly distributed variables. Fig. 3(a) gives the scatter plot of $t_1$ and $e_2$ used in this simulation and their marginal histograms. After that, we aimed to find variables $z_2$ and $e_1$, which are involved in the causal relation $x_2 \rightarrow x_1$, by making use of the transformation given in (6) and (7). Apparently, $z_2 = h(t_1) + e_2$. In order to find variable $e_1$, one needs to find $h_1(z_2)$. We first calculated $h_1(z_2)'$ according to (5). It is also given in Fig. 2(a), and the solution of $h_1$ with the initial condition $h_1(0) = 0$ is shown in Fig. 2(b). Variable $e_1$ is then found as $e_1 = t_1 + h_1(z_2)$. The scatter plot of $z_2$ and $e_1$ is shown in Fig. 3(b). Results of statistical independence tests, given in the left part of Table 3, confirm statistical independence between $t_1$ and $e_2$ and that between $z_2$ and $e_1$.

Analogously to Simulation I, we constructed observable data $(x_1, x_2)$ from $t_1$, $e_2$, and $h$ given above. To do that, we let $g_2(t_1) = t_1 + t_1^3$ and $f_2(h(t_1)+e_2) = \log(6 + h(t_1) + e_2)$, and $f_1$ was constructed as $f_1 = h \circ g_2^{-1}$. The nonlinear ICA-based method was then applied to the data $(x_1, x_2)$. As shown in the right part of Table 3, under both hypotheses $x_1 \rightarrow x_2$ and $x_2 \rightarrow x_1$, the assumed cause is independent from the estimate of the disturbance. Consequently, both causal directions can explain the data, and in this example the PNL causal model is not identifiable.

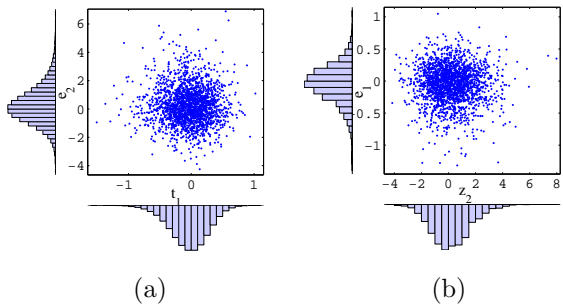

Figure 3: (a) Scatter plot of $t_1$ and $e_2$, which are independent variables with the densities given in Fig. 2(b). (b) Scatter plot of $z_2$ and $e_1$, which were obtained by making use of the transformation in (6) and (7).

Table 3: Result of independence tests in Simulation 2 at the significance level $\alpha = 0.05$. The independence hypothesis is accepted in all the four cases.

|  | $t_1 \& e_2$ | $z_2 \& e_1$ | $x_1 \& \hat{e}_2$ | $x_2 \& \hat{e}_1$ |
|---|---|---|---|---|
| Threshold | 0.5483 | 0.1561 | 0.5081 | 0.6912 |
| Statistic | 0.1641 | 0.0669 | 0.3011 | 0.4142 |

# 6   CONCLUSION

We have investigated the identifiability of the post-nonlinear (PNL) causal model in the two-variable case, and a practical identification method based on nonlinear function approximations and independence tests was discussed. Our results show that this model is generally identifiable, and consequently can be used to distinguish the cause from effect. All the particular situations in which this model is not identifiable were reported in Theorem 8, and some of them were verified and illustrated by simulations. For the situation with more than two variables, we showed that it is not necessary to apply the PNL causal model to all of the given variables directly, which becomes intractable as the variable number increase; instead, to find the whole causal structure, one can apply this causal model to the Markov equivalent class and test if the estimated disturbance is independent from the parents associated with the same variable.

## Acknowledgements

The authors would like to thank Patrik Hoyer for helpful comments and suggestions.

# APPENDIX: SOME PROOFS

**Proof of Theorem 1:** We prove this theorem using the linear separability of the logarithm of the joint density of independent variables, which states the fact that for a set of independent random variables whose joint density is twice differentiable, the Hessian of the logarithm of their density is diagonal everywhere (Lin, 1998). Since $g_2$ is invertible, the independence between $x_1$ and $e_2$ is equivalent to that between $x_1$ and $e_2$. Similarly, the independence between $x_2$ and $e_1$ is equivalent to that between $z_2$ and $e_1$. Combining the two causal models (2) and (3), one can see that the transformation from $(z_2, e_1)$ to $(t_1, e_2)$ is

$$t_1 = h_1(z_2) + e_1, \qquad (6)$$
$$e_2 = z_2 - h(t_1). \qquad (7)$$

Denote by $\mathbf{J}$ the Jacobian matrix of this transformation. One can see that $|\mathbf{J}| = 1$. Denote by $p_{(z_2, e_1)}$ the joint density of $(z_2, e_1)$. We then have $p_{t_1} \cdot p_{e_2} = p_{(z_2, e_1)}/|\mathbf{J}| = p_{(z_2, e_1)}$, so, $\log p_{(z_2, e_1)} = \eta_1(t_1) + \eta_2(e_2)$. One can find the (1,2)-th entry of the Hessian matrix of $\log p_{(z_2, e_1)}$ w.r.t. $(z_2, e_1)$: $\frac{\partial^2 \log p_{(z_2, e_1)}}{\partial e_1 \partial z_2} = \eta_1'' \frac{\partial t_1}{\partial z_2} - \eta_2' h' \frac{\partial e_2}{\partial z_2} - \eta_2' h'' \frac{\partial t_1}{\partial z_2} = \eta_1'' h_1' - \eta_2'' h' + \eta_2'' h'^2 h_1' - \eta_2' h'' h_1'$.

The independence between $z_2$ and $e_1$ implies $\frac{\partial^2 \log p_{(z_2, e_1)}}{\partial e_1 \partial z_2} = 0$ for every possible $(z_2, e_1)$. That is, $\eta_1'' h_1' - \eta_2'' h' + \eta_2'' h'^2 h_1' - \eta_2' h'' h_1' = 0$. From this equation one can see that $h_1' = 0$ implies $\eta_2'' h' = 0$. Consequently, the points which satisfy $\eta_2'' h' \neq 0$ also make $h_1' \neq 0$. For such points, dividing both sides of this equation by $h_1' \eta_2'' h'$ finally leads to (5). Furthermore, since $h_1$ is a functions of $z_2$ and does not depend on $e_1$, we have $\partial\left(\frac{1}{h_1'}\right) \big/ \partial e_1 = 0$. According to (5), we have $\partial\left(\frac{\eta_1'' + \eta_2'' h'^2 - \eta_2' h''}{\eta_2'' h'}\right) \big/ \partial e_1 = 0$, which gives $2\eta_2''^2 h'^2 h'' - \eta_2' \eta_2'' h' h''' + \eta_2' \eta_1''' h' - \eta_2' \eta_2'' h'^2 h'' + \eta_2' \eta_2'' h''^2 + \eta_1''' h'' h'^2 - \eta_2' \eta_1'' h'' = 0$. For the points satisfying $\eta_2'' h' \neq 0$, we divide both sides of the above equation by $\eta_2'' h'$. After some simplifications, (4) is obtained. ∎

**Proof of Lemma 6:** Gaussianity of $e_2$ implies that $\eta_2'' \equiv 0$ and that $\eta_2''$ is constant. (4) then reduces to

$$\eta_1''' - \frac{\eta_1'' h''}{h'} + 2\eta_2'' \cdot h' h''' = \eta_2' \cdot \left(h''' - \frac{h''^2}{h'}\right). \qquad (8)$$

Since the left-hand side does not depend on $e_2$ and $\eta_2'$ is a function of $e_2$, we have $h''' - \frac{h''^2}{h'} = 0$, which gives $h'''/h' - \frac{h''^2}{h'^2} = 0$. That is, $(h''/h')' = 0$, so $h''/h' = c_1$. If $c_1 = 0$, $h$ is linear, and $h'' = h''' = 0$. (8) then yields $\eta_1''' = 0$, meaning that $t_1$ is Gaussian.

Otherwise, $h' = \pm e^{c_1 t_2 + c_2}$. Moreover, the left-hand side of (8) must also be zero, which means $\frac{\eta_1'''}{h'} - \frac{\eta_1'' h''}{h'^2} +$

$2\eta_2'' h'' = 0$. By integration, it gives $\int \left(\frac{\eta_1'''}{h'} - \frac{\eta_1'' h''}{h'^2} + 2\eta_2'' h''\right) dt_1 = \eta_1''/h' + 2\eta_2'' h'' = c_3$, so $\eta_1'' = -2\eta_2'' h'^2 + c_3 h' = -2\eta_2'' e^{2c_1 t_1 + 2c_2} \pm c_3 e^{c_1 t_1 + c_2}$. Consequently $\eta_1 = \frac{-\eta_2'}{2c_1^2} e^{2c_1 t_1 + 2c_2} \pm \frac{c_3}{c_1^2} e^{c_1 t_1 + c_2} + c_4 t_1 + c_5$. Noting that $\eta_2''$ is a negative constant, we can see that $\eta_1 \to +\infty$ when $c_1 t_1 \to +\infty$. This contradicts Lemma 5. Thus, $c_1$ must be zero, $h$ is linear, and $t_1$ is Gaussian. ∎

**Proof of Lemma 7:** Note that $\eta_2''$ (as well as $\eta_1''$) is not constantly zero, as a density function of $e_2$ could not be proportional to $e^{c_1 e_2 + c_2}$. When $h$ is linear, $h'$ is constant and $h'' = h''' = 0$, and (4) becomes

$$\eta_1''' = -\frac{\eta_2'''}{\eta_2''} \cdot h' \eta_1'', \quad \text{i.e.,} \quad \frac{\eta_1'''}{\eta_1''} = -h' \cdot \frac{\eta_2'''}{\eta_2''}.$$

Since $\frac{\eta_1'''}{\eta_1''}$ and $\frac{\eta_2'''}{\eta_2''}$ depends only on $t_1$ and $e_2$, respectively, the equation above implies that both $\frac{\eta_1'''}{\eta_1''}$ and $\frac{\eta_2'''}{\eta_2''}$ are constants. Let $\frac{\eta_2'''}{\eta_2''} = c_1$, we have $\frac{\eta_1'''}{\eta_1''} = -h' c_1$. If $c_1 = 0$, clearly both $e_2$ and $t_1$ are Gaussian, and proposition (i) holds.

Otherwise, we have $\log |\eta_2''| = c_1 e_2 + c_2$. Consequently $\eta_2'' = \pm e^{c_1 e_2 + c_2}$ and $\eta_2 = \pm \frac{1}{c_1^2} e^{c_1 e_2 + c_2} + c_3 e_2 + c_4$. If $\eta_2 = \frac{1}{c_1^2} e^{c_1 e_2 + c_2} + c_3 e_2 + c_4$, clearly $\eta_2 \to +\infty$ when $\text{sgn}(c_1) \cdot e_2 \to +\infty$, which contradicts Lemma 5. One can verify that when $c_1 c_3 > 0$, the solution

$$\eta_2'' = -e^{c_1 e_2 + c_2}, \text{ or } \eta_2 = -\frac{1}{c_1^2} e^{c_1 e_2 + c_2} + c_3 e_2 + c_4 \quad (9)$$

corresponds to valid densities. Analogously we have

$$\eta_1'' = -e^{-h' c_1 t_1 + c_5}, \text{ or } \eta_1 = -\frac{1}{h'^2 c_1^2} e^{-h' c_1 t_1 + c_5} + c_6 t_1 + c_7. \quad (10)$$

Thus the densities of $t_1$ and $e_2$ are both log-mix-lin-exp.

Combining (5), (9) and (10), and recalling that $h$ is linear, we have

$$h_1' = \frac{\eta_2'' h'}{\eta_1'' + \eta_2'' h'^2 - \eta_2' h''} = \frac{1}{\eta_1''/(\eta_2'' h') + h'}$$
$$= \frac{1}{e^{-c_1 z_2 + c_5 - c_2}/h' + h'}. \qquad (11)$$

One can see that $h_1' \to 0$, as $\text{sgn}(c_1) \cdot z_2 \to +\infty$, and as $\text{sgn}(c_1) \cdot z_2 \to -\infty$, $h_1' \to h'$, which is a constant. Therefore Lemma 7 is true. ∎

**Outline of Proof of Theorem 8:** Eq. (4) can be re-written as a bilinear functional equation (Polyanin & Zaitsev, 2004) of the form:

$$\Phi_1(t_1)\Psi_1(e_2) + \Phi_2(t_1)\Psi_2(e_2)$$
$$+ \Phi_3(t_1)\Psi_3(e_2) + \Phi_4(t_1)\Psi_4(e_2) = 0, \quad (12)$$

where

$$
\begin{aligned}
\Phi_1(t_1) &= \eta_1''' - \tfrac{\eta_1'' h''}{h'}, & \Phi_2(t_1) &= h''' - \tfrac{h''^2}{h'}, \\
\Phi_3(t_1) &= h'h'', & \Phi_4(t_1) &= h'\eta_1'', \\
\Psi_1(e_2) &= -1, & \Psi_2(e_2) &= \eta_2', \\
\Psi_3(e_2) &= \tfrac{\eta_2'\eta_2'''}{\eta_2'^2} - 2\eta_2'', & \Psi_4(e_2) &= -\tfrac{\eta_2'''}{\eta_2'}.
\end{aligned} \tag{13}
$$

Note that functionals $\Phi_i(t_1)$ and $\Psi_i(e_2)$ depend only on $t_1$ and $e_2$, respectively. We then find all possible situations in which (12) holds.

Clearly $\Phi_4(t_1)$, $\Psi_1(e_2)$, and $\Psi_2(e_2)$ in (13) are not constantly zero. We first consider some simple cases of the solutions to (12), with $\Phi_3(t_1) \equiv 0$, $\Psi_4(e_2) \equiv 0$, or $\Phi_2(t_1) \equiv 0$. These cases either have no valid solutions for $\eta_1$, $\eta_2$, and $h$, or the solutions are covered by the situations in Table 1.

When none of the functions mentioned above is constantly zero, it can be shown that the functional equation of the form (12) has three solutions, which are (Polyanin & Zaitsev, 2004)

Solution 1:

$$
\begin{aligned}
\Phi_1 &= A_1\Phi_3 + A_2\Phi_4, & \Phi_2 &= A_3\Phi_3 + A_4\Phi_4, \\
\Psi_3 &= -A_1\Psi_1 - A_3\Psi_2, & \Psi_4 &= -A_2\Psi_1 - A_4\Psi_2,
\end{aligned} \tag{14}
$$

Solution 2:

$$
\begin{aligned}
\Phi_1 &= B_1\Phi_3, & \Phi_2 &= B_2\Phi_3, & \Phi_4 &= B_3\Phi_3, \\
\Psi_3 &= -B_1\Psi_1 - B_2\Psi_2 - B_3\Psi_4,
\end{aligned} \tag{15}
$$

Solution 3:

$$
\begin{aligned}
\Psi_2 &= C_1\Psi_1, & \Psi_3 &= C_2\Psi_1, & \Psi_4 &= C_3\Psi_1, \\
\Phi_1 &= -C_1\Phi_2 - C_2\Phi_3 - C_3\Phi_4,
\end{aligned} \tag{16}
$$

where $A_i$, $B_i$, and $C_i$ are arbitrary constants. Each possible solution given above is an over-determined system of ordinary differential equations (ODE's). In most cases, solutions of $\eta_1$, $\eta_2$, and $h$ to the system, which also satisfy the properties given in Lemma 5, can be found in closed form. In the remaining cases, we use the way of phase portrait (Braun, 1993) to analyze the behavior of the solutions. We consider all possible cases, and the solutions are always covered by the five situations enumerated in Table 1. ∎

**Proof of Corollary 10:** First, consider the case where $f_1' \neq 0$ at every point or it is zero at only some discrete points. As $h = f_1 \circ g_2$ and $g_2$ is invertible, $h' = 0$ holds at most at some discrete points. All the situations in which the PNL causal model is not identifiable are given in in Table 1. One can see that in all these situations, $h = f_1 \circ g_2$ is invertible. When $f_1$ is not invertible, no matter what $g_2$ is, it is impossible to make $h = f_1 \circ g_2$ invertible, i.e., none of the conditions in Table 1 holds. Consequently, in this case the causal direction between $x_1$ and $x_2$ is uniquely identified.

Next, consider the case where $f' \equiv 0$ on the domain $\mathbb{D}_0 \in \mathbb{R}$. Let $C$ be a point between $\mathbb{D}_0$ and the domain $\mathbb{D}_n$ on which $f' = 0$ holds at most at some discrete points. Without loss of generality, we assume that $\mathbb{D}_n$ is on the right side of $\mathbb{D}_0$. As $h = f_1 \circ g_2$, we have $h(t_1 = C^-)' = 0$. Suppose that both $x_1 \to x_2$ and $x_2 \to x_1$ can explain the data. In all the situations listed in Table 1, $h' \neq 0$. Hence, $h(t_1 = C^+)' \neq 0$. As $h(t_1 = C^+)' \neq h(t_1 = C^-)$, $h$ is not differentiable at $t_1 = C$, which causes a contradiction. So in this case the causal direction between $x_1$ and $x_2$ given by the PNL causal model is also unique. ∎

**Proof of Theorem 11:** The necessity part is obvious, and below we prove the sufficiency part, which states that disturbances $e_1, ..., e_n$ are mutually independent, or equivalently, $x_1, ..., x_n$ follow the PNL causal model represented by $\mathcal{G}$, if the causal Markov condition holds and the disturbance $e_i$ in the variable $x_i$ is independent of $pa_i$. Let $z_i \triangleq f_{i,2}^{-1}(x_i)$. As the causal relations are acyclic, we can arrange $x_i$ in an order such that no later variable causes any earlier one. Without loss of generality, we assume that $(x_1, x_2, ..., x_n)$ is one of such orders. For any $i = 1, ..., n$, since $x_i = f_{i,2}(z_i)$, we have $p(x_i|pa_i) = p(z_i|pa_i)/|f_{i,2}'(z_i)|$. Therefore,

$$
\begin{aligned}
H(e_i) &\geq H(e_i|pa_i) \tag{17} \\
&= H(z_i|pa_i) = -E\{\log p(z_i|pa_i)\} \\
&= -E\{\log p(x_i|pa_i)\} - E\{\log|f_{i,2}'(z_i)|\} \\
&= H(x_i|pa_i) - E\{\log|f_{i,2}'(z_i)|\} \\
&\geq H(x_i|x_1,...x_{i-1}) - E\{\log|f_{i,2}'(z_i)|\}, \tag{18}
\end{aligned}
$$

where $H(\cdot)$ denotes the entropy, the equality in (17) holds if and only if $e_i$ is independent from $pa_i$, and the equality in (18) holds if and only if the causal Markov condition holds, i.e., elements of $\{x_k|x_k \notin pa_i, 1 \leq k \leq i-1\}$ are independent of $x_i$ given $pa_i$. Taking the summation of (18) over i gives

$$
\begin{aligned}
&\sum_i H(e_i) \\
&\geq \sum_i H(x_i|x_1,...,x_{i-1}) - \sum_i E\{\log|f_{i,2}'(z_i)|\} \\
&= H(x_1,...,x_n) - \sum_i E\{\log|f_{i,2}'(z_i)|\}. \tag{19}
\end{aligned}
$$

In addition, since $e_i$ does not depend on $x_j$ ($j > i$), the Jacobian matrix of the transformation from $(x_1, ...x_n)$ to $(e_1, ...e_n)$ is lower-triangular, with the $(i,i)$th entry being $1/f_{i,2}'(z_i)$. Consequently, the absolute value of the determinant of this Jacobian matrix is $|\mathbf{J}| = [\Pi_i f_{i,2}'(z_i)]^{-1}$. Recalling (19), when the causal Markov condition holds and $e_i$ is independent of $pa_i$, the mutual information of $e_1, ...e_n$ is then $I(e_1, ..., e_n) = \sum_i H(e_i) - H(e_1, ..., e_n) = \sum_i H(e_i) - [H(x_1, ..., x_n) + \log|\mathbf{J}|] = \sum_i H(e_i) - H(x_1, ..., x_n) + \sum_i \log|f_{i,2}'(v_i)| = 0$. That is, $e_1, ..., e_n$ are mutually independent. ∎